\documentclass{article}

\usepackage[preprint]{neurips_2026}

\usepackage[utf8]{inputenc} 
\usepackage[T1]{fontenc}    
\usepackage{hyperref}       
\usepackage{url}            
\usepackage{booktabs}       
\usepackage{amsfonts}       
\usepackage{nicefrac}       
\usepackage{microtype}      
\usepackage{xcolor}         

\usepackage[breakable,skins]{tcolorbox} 
\usepackage{amsmath} 
\usepackage{graphicx} 

\title{\underline{G}ener\underline{a}lization or \underline{M}emorization? \underline{B}r\underline{i}ttleness \underline{T}esting for Chess-Trained Language Models}

\author{%
  Ethan Tang \\
  School of Computing and Augmented Intelligence \\
  Arizona State University \\
  Tempe, AZ 85281 \\
  \texttt{ejtang@asu.edu} \\
}

\begin{document}

\maketitle

\begin{abstract}

Recent work has fine-tuned language models on chess data and reported high benchmark scores as evidence that the resulting models can understand the rules of chess, play full chess games at a professional level, or generate human-readable explanations grounded in expert knowledge. We train KinGPT, a 25M-parameter character-level language model trained only on (position, best-move) pairs, who exceeds 3B-parameter ChessGPT on a 600-puzzle mate-in-N suite and 4B-parameter C1-4B over a 20-theme puzzle benchmark. We examine several claims made in existing literature regarding chess-trained language models and assert that their impressive benchmark performance is largely explained by pattern-matching. We also demonstrate how LLM-Modulo, a verifier-in-the-loop framework, raises RedPajama 3B's best move accuracy from 1.2\% to 21.2\% and move generation validity from 19.3\% to 95.3\% on mate-in-N chess puzzles, comparable to gains achieved from ChessGPT's fine-tuning on chess-specific web corpora at a fraction of the cost. Our results illustrate how pairing a general LLM with an external verifier offers a more flexible alternative to directly training on synthetic data for well-defined domains. We open source all training/evaluation code, datasets, puzzle samples, and KinGPT model checkpoints for reproducibility.

\end{abstract}

\section{Introduction}

Chess has been commonly referred to as the drosophila of artificial intelligence~\cite{mccarthy1990chessdrosophila} due to its rules providing a simple, well-defined domain for testing. With the advent of GPT-3 in 2020 \cite{brown2020language}, followed by the introduction of RLHF to capture user preferences \cite{ouyang2022training} and CoT/RLVR to elicit reasoning-like behavior \cite{wei2022chain,DeepSeekAI2025DeepSeekR1IR}, chess has become an attractive testbed for modeling the capabilities of language models.

Starting in 2023, several attempts to distill general chess ability into language models have garnered significant interest, claiming language models are capable of learning the basic rules of chess and obtaining the ill-defined property of ``chess understanding''. We demonstrate that KinGPT, a 25M-parameter language model with no capability beyond the domain of chess puzzles, outperforms recent open-sourced chess-trained language models at a large sample of beginner-level puzzles, showcasing that strong performance in chess benchmarks via pattern-matching does not generalize to a deeper understanding of the game. We also discuss several methodological gaps made in the current literature and propose KinGPT as a baseline model for evaluating chess-trained language models on the domain of chess puzzles. Finally, we highlight the efficacy of the LLM-Modulo framework applied to the chess puzzle domain, illustrating how verifier-in-the-loop methods provide substantial gains to both move generation validity and accuracy for general language models and yield comparable gains for specialized language models.

\section{Related work}

\cite{feng2023chessgpt} fine-tuned the RedPajama-3B-base model \cite{together2023redpajama} on a large scale web corpora of human/engine chess games, puzzles, annotations, and chess websites. Evaluating their model on several benchmarks in the chess domain, ChessGPT achieved \textasciitilde71.4\% accuracy on a set of checkmate-in-1 puzzles, providing a prime example of ``the model’s capacity to comprehend and apply the rules of chess'' (p.~9).

\cite{zhang2025complete} fine-tuned the OpenLLaMa-3B-V1 model \cite{openlm2023openllama,touvron2023llama} on a large corpus of (position, best move) pairs from Stockfish-labeled human and self-play games, achieving close to perfect (99.11\%) accuracy for legal move generation using pass@1 sampling. Playing full chess games is a much harder task than solving individual chess puzzles; Zhang et al. asserted that they ``[were] the only ones using a large language model for chess that can complete full games'' (p.~1). Their model, ChessLLM, achieved a performance rating of 1530 Elo\footnote{This does not match \cite{zhang2025complete}'s Elo calculations, which use a game-by-game update with an undisclosed K-factor instead of true performance rating \cite{true-perf-rating,chessLLM-perf-rating}.} using pass@10 sampling in 100-game matches versus Stockfish Levels 0-2, further strengthening their claim that ChessLLM ``[obtained] the extrapolation and combat capabilities of chess games'' (p.~6).

The most recent work by Z. \cite{tang2026groundedchessreasoninglanguage} introduced the Master Distillation framework and demonstrated how it could be leveraged to ``inject expert-level knowledge into compact models for under-optimized domains'' (p.~1). Z. Tang et al. fine-tuned the Qwen3-4B-Instruct-2507 model \cite{qwen3technicalreport} on large samples of chess puzzles from the Lichess puzzle database, followed by additional post-training via RLVR based on a binary reward of Stockfish-labeled best moves. Z. Tang et al.'s model, C1-4B, achieved an average first-move accuracy of 48.1\% on a varied set of chess puzzle themes using pass@1 sampling along with ``exhibit[ing] remarkable token efficiency, generating solutions in roughly two orders of magnitude fewer tokens than both proprietary and open-source baselines'' (p.~6). Additionally, Z. Tang et al. hypothesized how their C1-4B model could ``generate human-readable explanations grounded in expert knowledge'' (p.~9) to provide a bridge between model proficiency and human understanding of chess via intermediate reasoning traces.

\section{Our contribution}

Both fine-tuning a language model and post-training on high-quality synthetic chess data (e.g. position + best move pairs) can be seen as methods for compiling an external verifier's signal (e.g. a chess engine) into a language model's weights. The clear limitation to both methods is a lack of validity or correctness guarantees at inference time, as their additional training on synthetic data only increases the probability of success when asked to solve any particular chess-related task.

We train a small language model, KinGPT, on a corpus consisting purely of chess (position, best move) pairs to provide a realistic benchmark on chess puzzle performance for future work investigating chess-trained language models. Using KinGPT's performance as a baseline, we examine several claims made by \cite{feng2023chessgpt,zhang2025complete,tang2026groundedchessreasoninglanguage} on the capabilities of chess-trained language models. Additionally, we demonstrate that the LLM-Modulo framework defined by \cite{kambhampati2024position}, consisting of pairing a language model with an external verifier, can be applied effectively to the chess domain. Finally, we discuss the significance and implications of language models' performance on chess-related benchmarks and how chess-trained language models may not neatly fit the capabilities they have been ascribed.

\section{KinGPT}
\label{KinGPT}

We trained 3 variants of KinGPT, a 25M-parameter language model utilizing character-level tokenization, using code forked from the nanoGPT repository created by \cite{karpathy-nanoGPT}. Each variant was trained on a corpus consisting solely of chess (position, best move) pairs, with the only major difference between variants being the contents of their training corpus.

\clearpage

\textbf{KinGPT-Woodpecker} (named after \cite{Smith2024woodpecker}) was trained on a corpus of 13,341,057 unique puzzle positions from the Lichess puzzle database (\cite{lichess-puzzle-db}) for 500 billion tokens (characters). All puzzle positions used in training were filtered to ensure no FEN-level position overlap occurred with the evaluation set of puzzles used later in Section \ref{sec:results}.

\textbf{KinGPT-Beaver} was trained on a corpus of 54,681 unique positions extracted from 1050 Stockfish 18 self-play games for 25 billion tokens. Self-play games were generated between a base version of Stockfish 18 (depth=15, time=10s) and weaker variants utilizing the UCI Option ``Skill Level'' (\cite{sf-wiki-skilllevel}) at the same depth and think time.

\textbf{KinGPT-Chimera} was trained on a combined corpus of 13,395,738 (position, best move) pairs from the Woodpecker and Beaver variants for 500 billion tokens. 

For more information about training configuration or inference for KinGPT models, please refer to \url{https://github.com/ethanjtang/KinGPT}.

\section{Dataset}

To form the training corpus of KinGPT-Woodpecker, we performed a theme-wide split of puzzles in the Lichess puzzle database (\cite{lichess-puzzle-db}), withholding 1000 puzzles from each theme for model evaluations. In total, 72625 total unique validation puzzles were sampled out of 74 puzzle themes (after de-duplicating puzzles assigned multiple themes). The full training and validation split of puzzles, including samples for each theme, can be found at \url{https://huggingface.co/datasets/ethanjtang/GAMBIT-lichess-puzzle-positions}. 

To form the training corpus of KinGPT-Beaver, we paired a base version of Stockfish 18 (SF18 Base) against weaker variants of itself using the UCI Option ``Skill Level''. SF18 Base played 50 games (25 White, 25 Black) against each model variant (Levels 0-20) at depth=15 and time=10s for each move. No opening books were used in self-play games, resulting in some overlap (\textasciitilde9\%) in positions between games. The full set of Stockfish 18 self-play games can be found at \url{https://huggingface.co/datasets/ethanjtang/GAMBIT-stockfish18-selfplay}

\section{Evaluations}

\subsection{Models evaluated}
\label{models-evaluated}

For evaluating all language models, we extracted a random sample of $n=100$ puzzles with the following themes of ``mateIn1'', ``mateIn2'', and ``mateIn3''. We tested the following models on the full $n=600$ suite of puzzle positions ($n=100$ for mate-in-1, $n=200$ for mate-in-2, $n=300$ for mate-in-3), evaluating the number of legal moves made (Eq. \ref{sanity}), number of accurate moves made (Eq. \ref{overall-acc}), and puzzle-wide accuracy (Eq. \ref{puzzle-acc}) for each model.

\begin{itemize}

\item{\textbf{OpenLLaMa 3B V1} - base model of \cite{zhang2025complete}'s model ChessLLM before finetuning on chess synthetic data}
\item{\textbf{RedPajama 3B Base} - base model of \cite{feng2023chessgpt}'s model ChessGPT before finetuning on web-scale chess data}
\item{\textbf{ChessGPT} - both Base and Chat variants trained by \cite{feng2023chessgpt}}
\item{\textbf{KinGPT} - all 3 variants of the 25M-parameter language model (Section \ref{KinGPT})}

\end{itemize}

We also performed a theme-wide performance comparison of KinGPT-Chimera against \cite{tang2026groundedchessreasoninglanguage}'s C1-4B model, measuring first-move accuracy on a random sample of $n=100$ puzzles from the same distribution of $n=20$ themes C1-4B was evaluated on. We note that our random sample of $n=100$ puzzles is not an exact match\footnote{As of 5/16/2026, puzzle samples and model checkpoints for C1-4B have not been published to their public repository at \url{https://github.com/CSSLab/C1}}, but we open source our random sample for reproducibility at \url{https://github.com/ethanjtang/GAMBIT/tree/main/sample_puzzles/chimera-vs-c1-samples}.

\subsection{Inference types}

\subsubsection{Overview}

For the $n=300$ sample of mate-in-N puzzles, we evaluate all models using normal, cheating, pass@10, and LLM-Modulo (modulo) style prompting as follows:

\begin{itemize}
\item{\textbf{normal} - Model is queried a single time (pass@1) for the best move in the given puzzle position.}
\item{\textbf{cheating} - Model is queried a single time for the best move in the given puzzle position, with a string prepended to the prompt stating the current evaluation of the position (following the definition in \cite{feng2023chessgpt}). For mate-in-N puzzles, this means informing the model of the number of moves to deliver checkmate for the side-to-move.}
\item{\textbf{pass@10} - Model is queried up to $K=10$ times (temperature=0.7) for the best move in a given puzzle position.}
\item{\textbf{modulo} - Model is queried up to $K=10$ times (temperature=0.7) for the best move in a given puzzle position. Two hard critics gate model output, with Critic \#1 checking move validity (Is the generated move legal in the given position?) and Critic \#2 checking move accuracy (Does the generated move improve the evaluation of the original position?). Model output must pass through both critics before being accepted. If either critic rejects a model's generated move, it is re-prompted with feedback on why the generated move was rejected. Appendix \ref{modulo_critic1} and \ref{modulo_critic2} contain examples of model re-prompting with feedback from Critic \#1 and \#2.}
\end{itemize}

\subsubsection{LLM-Modulo loop implementation}

Preliminary tests with the OpenLLaMa model yielded several issues with modulo-style prompting which were addressed with the following optimizations. We filter ``Model:'' and ``User:'' tags from model output to remove model-hallucinated role tags. We also re-prompt with additional feedback only when the model makes a different move from its previously parsed output, preventing prompt context from increasing due to feedback for repeated identical moves. Finally, we check that the model is producing coherent responses (Eq. \ref{sanity}) throughout the re-prompting loop. Upon detecting 3 consecutive parse failures, a model's context window is reset by re-prompting it with the original prompt with all feedback removed. This reset addresses scenarios where the iterative feedback provided by Critics \#1 and \#2 is no longer effective, as the model continues to repeatedly output invalid moves. For additional reference for modulo-style prompting, please refer to source code at \url{https://github.com/ethanjtang/GAMBIT/blob/main/eval_models_on_puzzles/eval_all_models_modulo.py}.

\subsection{Model accuracy}

For each language model, we evaluate both puzzle-wide and position-wide accuracy. Position-wide (overall) accuracy measures how frequently a model identifies the best move across all tested puzzle positions. 

\begin{equation}
\label{overall-acc}
\text{Overall Accuracy} = \frac{\text{\# of correct responses}}{\text{total \# of positions}}
\end{equation}

Puzzle-wide accuracy is a harder metric than position-wide accuracy, as a puzzle is considered solved only if the model finds the best move for all puzzle positions (e.g. A mate-in-3 puzzle requires 3 positions to be solved correctly).

\begin{equation}
\label{puzzle-acc}
\text{Puzzle Accuracy} = \frac{\text{\# of puzzles solved correctly}}{\text{total \# of puzzles}}
\end{equation}

For normal and cheating-style prompting, a model's response is considered correct for any given position if it matches the Lichess puzzle's ground truth solution or is considered by Stockfish to be an ``alternatively good'' solution. A move is considered to be ``alternatively good'' if it improves on the evaluation of the given position. For mate-in-N puzzles with a position evaluated at mate-in-N, an alternative solution would yield a position with the evaluation of mate-in-(N-1).

For pass@10 and modulo-style prompting, a model's response is considered correct if any of the $K=10$ responses would be considered correct under the previous definition. Puzzle-wide accuracy is calculated similarly across all prompting types, with a puzzle considered to be solved correctly if a model was able to find a correct solution for every puzzle position.

\subsection{Model sanity}

For each language model, we also evaluate sanity, which is inversely proportional to the number of invalid model responses (move parse failures). Sanity measures the frequency models output a legal/valid move across all tested puzzle positions.

\begin{equation}
\label{sanity}
\text{Sanity} = 1 - \frac{\text{\# of invalid parses}}{\text{total \# of positions}}
\end{equation}

For normal and cheating-style prompting, a model's response is treated as a parse failure if we are unable to parse a valid UCI/SAN string (\cite{cpw-chess-notation}) representing a valid chess move in the position. For pass@10 and modulo-style prompting, a model's response is treated as a parse failure only if all $K=10$ responses result in parse failures.

\section{Results}
\label{sec:results}

\subsection{Stockfish vs. KinGPT variants}
\label{results-sf-KinGPT}

\begin{table}[ht]
  \caption{Model accuracy for Stockfish 18 and KinGPT on $n=300$ mate-in-N puzzles. Stockfish 18 Base at depth=20 acts as our ground truth. All reported results use a 99\% Wilson score interval.}
  \label{sf18-KinGPT-results}
  \centering
  \begin{tabular}{llrr}
    \toprule
    Model & Inference & Puzzle Accuracy (\%) & Position Accuracy (\%) \\
    \midrule
    SF 18 Base & depth=20 & -- & -- \\
    SF 18 Base & time=0.05s & $98.9 \pm 1.1$ & $99.5 \pm 0.5$ \\
    SF 18 Level 0 & depth=20 & $63.7 \pm 7.1$ & $79.3 \pm 4.2$ \\
    \midrule
    Woodpecker & normal & $71.9 \pm 6.6$ & $81.7 \pm 4.0$ \\
    Beaver & normal & $2.1 \pm 1.8$ & $2.2 \pm 1.4$ \\
    Chimera & normal & $74.5 \pm 6.4$ & $84.6 \pm 3.8$ \\
    \bottomrule
  \end{tabular}
\end{table}

Our ground truth, Stockfish 18 Base, gives the correct move across all puzzle positions, even when reducing inference speed down to 0.05 seconds. This validates that the many external verifiers (e.g. chess engines) available for chess significantly outperform all open-sourced chess-trained language models including KinGPT.

Among the KinGPT variants, Chimera performs the best followed closely by Woodpecker. Incorporating additional positions from self-play games to a corpus mainly consisting of puzzle positions (Chimera -- 84.6\%) boosts performance a negligible amount compared to training on only puzzle positions (Woodpecker -- 81.7\%). In contrast, Beaver, which was trained on only position + best move pairs from Stockfish self-play games, performs poorly (2.2\%) relative to the other two variants.

Beaver's failure does not directly impugn \cite{zhang2025complete}'s ChessLLM, which trained on a substantially larger scale of Stockfish-labeled (position, best move) pairs. However, it raises the possibility that puzzle-solving competence does not emerge cheaply from training on (position, best move) pairs extracted from game data, despite both tasks sharing the same underlying objective of finding the best move in any given position. We cannot test this directly: \cite{zhang2025complete} provide insufficient details for reproducing their model and did not release their training code, dataset, or model checkpoints.

\subsection{General models vs. ChessGPT}
\label{results-general-chessgpt}

\begin{table}[ht]
  \caption{Model accuracy for OpenLLaMa 3B, RedPajama 3B, and ChessGPT on $n=300$ mate-in-N puzzles. Modulo-style prompting outperforms pass@10 for general models, while chess-trained language models benefit more from pass@10 than modulo. All reported results use a 99\% Wilson interval.}
  \label{pajama-chessgpt-results}
  \centering
  \begin{tabular}{clrr}
    \toprule
    Model & Inference & Puzzle Accuracy (\%) & Position Accuracy (\%) \\
    \midrule
    Open LLaMa 3B V1 & normal & $1.1 \pm 1.1$ & $0.7 \pm 0.7$ \\
    -- & cheating & $2.7 \pm 2.2$ & $2.7 \pm 1.6$ \\
    -- & pass@10 & $2.1 \pm 1.8$ & $3.8 \pm 1.9$ \\
    -- & modulo & $3.7 \pm 2.6$ & $12.1 \pm 3.4$ \\
    \midrule
    Red Pajama 3B Base & normal & $1.1 \pm 1.1$ & $1.2 \pm 1.0$ \\
    -- & cheating & $1.1 \pm 1.1$ & $0.5 \pm 0.5$ \\
    -- & pass@10 & $1.4 \pm 1.4$ & $2.0 \pm 1.4$ \\
    -- & modulo & $10.9 \pm 4.5$ & $21.2 \pm 4.3$ \\
    \midrule
    ChessGPT-Base & normal & $16.1 \pm 5.4$ & $27.9 \pm 4.7$ \\
    -- & cheating & $16.7 \pm 5.4$ & $30.5 \pm 4.8$ \\
    -- & pass@10 & $38.6 \pm 7.2$ & $58.7 \pm 5.1$ \\
    -- & modulo & $18.7 \pm 5.7$ & $33.8 \pm 4.9$ \\
    \midrule
    ChessGPT-Chat & normal & $10.9 \pm 4.5$ & $21.3 \pm 4.3$ \\
    -- & cheating & $13.1 \pm 4.9$ & $25.8 \pm 4.6$ \\
    -- & pass@10 & $21.0 \pm 6.0$ & $38.0 \pm 5.1$ \\
    -- & modulo & $14.5 \pm 5.1$ & $29.6 \pm 4.8$ \\
    \bottomrule
  \end{tabular}
\end{table}

OpenLLaMa and RedPajama exhibit very poor puzzle-wide accuracy for both normal and cheating-style prompts, resulting in pass@10 results also being similarly low. In contrast, under modulo-style prompting, there is a statistically significant increase in both puzzle (1.4 $\to$ 10.9\%) and position-wide accuracy (2.0 $\to$ 21.2\%) for RedPajama and position-wide accuracy (3.8 $\to$ 12.1\%) for OpenLLaMa. This provides strong evidence that the LLM-Modulo framework boosts performance for general models on out-of-distribution domains, guiding model output via re-prompting with feedback from a domain-specific verifier.

In contrast, modulo-style prompting for ChessGPT variants boosts position and puzzle-wide accuracy to a lesser extent compared to pass@10. ChessGPT-Base achieves statistically significantly higher position-wide accuracy when comparing between modulo and pass@10-style prompting (33.8 vs. 58.7\%). This suggests that the LLM-Modulo framework is less effective for models which have already compiled parts of the verifier signal into their weights. 

Additionally, we note that our results do not match \cite{feng2023chessgpt}'s reported improvement of 44.9\% (26.5 $\to$ 71.4\%) on position-wide accuracy (Table~6) between normal and cheating-style prompting for ChessGPT-Base. On our $n=300$ sample of mate-in-N puzzles, we report only a statistically insignificant improvement of 2.6\% (27.9 $\to$ 30.5\%) for ChessGPT-Base and 4.5\% (21.3 $\to$ 25.8\%) for ChessGPT-Chat between normal and cheating-style prompting. We hypothesize that chess-trained language models may be adversely sensitive to puzzle position difficulty or prompt format, possessing brittle capabilities in the chess domain depending on the benchmark tested.

\subsection{Overall model sanity}
\label{results-sanity}

Modulo-style inference vastly improves sanity for general language models, with OpenLLaMa improving from 51.5\% to 81.3\% and RedPajama improving from 50.8\% to 95.3\% for pass@10 vs. modulo-style prompting. Meanwhile, for the specialized ChessGPT models, both pass@10 and modulo-style prompting appear to yield a similar improvement to model sanity for both Base (99.5\% vs. 98.3\%) and Chat (94.8\% vs. 97.3\%) variants.

\clearpage

\begin{table}[ht]
  \caption{Overall sanity for all models on $n=300$ mate-in-N puzzles. Stockfish and KinGPT variants consistently provide legal moves using pass@1 inference. Modulo-style prompting yields significant improvements to sanity over pass@10 for general models, and equivalent improvements to sanity for chess-trained language models.}
  \label{overall-sanity-results}
  \centering
  \begin{tabular}{llr}
    \toprule
    Model & Inference & Sanity (\%) \\
    \midrule
    Stockfish 18 & all & 100.0 \\
    KinGPT-Woodpecker & normal & 98.5 \\
    KinGPT-Beaver & normal & 28.3 \\
    KinGPT-Chimera & normal & 99.5 \\
    Open LLaMa 3B & normal & 5.5 \\
    Open LLaMa 3B & cheating & 20.0 \\
    Open LLaMa 3B & pass@10 & 51.5 \\
    Open LLaMa 3B & modulo & 81.3 \\
    Red Pajama 3B & normal & 19.3 \\
    Red Pajama 3B & cheating & 3.1 \\
    Red Pajama 3B & pass@10 & 50.8 \\
    Red Pajama 3B & modulo & 95.3 \\
    ChessGPT-Base & normal & 83.7 \\
    ChessGPT-Base & cheating & 81.0 \\
    ChessGPT-Base & pass@10 & 99.5 \\
    ChessGPT-Base & modulo & 98.3 \\
    ChessGPT-Chat & normal & 76.3 \\ 
    ChessGPT-Chat & cheating & 66.3 \\
    ChessGPT-Chat & pass@10 & 94.8 \\
    ChessGPT-Chat & modulo & 97.3 \\
    \bottomrule
  \end{tabular}
\end{table}

This showcases how the LLM-Modulo framework is more generalizable and less costly to implement for well-defined domains. Modulo-style prompting achieves comparable or improved sanity for general language models when compared against traditional methods of finetuning on web-scale domain-specific corpora. Another advantage of the LLM-Modulo framework over training on synthetic data is its flexibility, as both the verifier (Stockfish) and domain (chess) can be easily exchanged for any other well-defined domains with easily-accessible external verifiers (e.g. math, coding, etc.).

Both inference style and the distribution of positions in a model's training corpus measurably affect a language model's sanity, providing additional empirical evidence for their brittleness in the chess domain. KinGPT-Woodpecker and Chimera (98.5/99.5\%) outperform Beaver (28.3\%), demonstrating that the underlying distribution of positions is a dominant factor regarding legal move generation ability for chess-trained language models. In direct contrast, the respective improvement for OpenLLaMa (5.5 $\to$ 20.0\%) and degradation for RedPajama (19.3 $\to$ 3.1\%) between normal and cheating-style prompting is surprising, given that both models are of similar scale (3B-parameters) and utilize the same corpus for training (\cite{together2023redpajama}). This demonstrates how language models can be idiosyncratically sensitive to prompt format due to differences in model architecture, tokenizer, and/or training configuration.

\subsection{KinGPT-Chimera vs. C1-4B}
\label{results-KinGPT-c14b}

We performed a theme-wide comparison of model performance between KinGPT-Chimera and \cite{tang2026groundedchessreasoninglanguage}'s C1-4B model. As stated earlier in Section \ref{models-evaluated}, a direct one-to-one comparison is not currently possible. KinGPT-Chimera was tested on a random $n=100$ sample for each of the same $n=20$ puzzle themes as C1-4B, measuring first-move accuracy and comparing against reported first-move accuracy for C1-4B on \cite{tang2026groundedchessreasoninglanguage}'s puzzle sample. Checking for ``alternatively good'' solutions was only performed on puzzle positions with mate-in-N evaluations.

\begin{table}[ht]
  \caption{Model accuracy (\%) for a theme-wide comparison between KinGPT-Chimera and C1-4B. KinGPT-Chimera exhibits comparable performance to C1-4B for first-move accuracy across all chosen puzzle themes. All reported results use a 99\% Wilson interval.}
  \label{KinGPT-vs-c14b-results}
  \centering
  \begin{tabular}{lll}
    \toprule
    Theme & KinGPT & C1-4B \\
    \midrule
    advancedPawn & $60.3 \pm 12.2$ & $61.1 \pm 22.2$ \\
    attraction & $63.1 \pm 12.0$ & $61.1 \pm 22.2$ \\
    backRankMate & $91.3 \pm 6.5$ & $76.9 \pm 18.2$ \\
    capturingDefender & $64.1 \pm 11.9$ & $54.7 \pm 22.8$ \\
    defensiveMove & $60.3 \pm 12.2$ & $57.9 \pm 22.5$ \\
    deflection & $71.6 \pm 11.2$ & $38.9 \pm 22.2$ \\
    discoveredAttack & $60.3 \pm 12.2$ & $45.3 \pm 22.8$ \\
    doubleCheck & $72.5 \pm 11.0$ &  $51.6 \pm 22.9$ \\
    fork & $65.0 \pm 11.9$ &  $38.9 \pm 22.2$ \\
    hangingPiece & $79.1 \pm 10.0$ & $61.1 \pm 22.2$ \\
    mateIn1 & $84.7 \pm 8.7$ & $61.1 \pm 22.2$ \\
    mateIn2 & $82.8 \pm 9.2$ & $54.7 \pm 22.8$ \\
    pin & $61.3 \pm 12.1$ & $32.6 \pm 21.1$ \\
    promotion & $69.7 \pm 11.4$ & $51.6 \pm 22.9$ \\
    queensideAttack & $68.8 \pm 11.5$ & $64.2 \pm 21.7$ \\
    sacrifice & $55.6 \pm 12.4$ & $57.9 \pm 22.5$ \\
    skewer & $73.4 \pm 10.9$ & $51.6 \pm 22.9$ \\
    trappedPiece & $48.1 \pm 12.5$ & $13.6 \pm 13.2$ \\
    xRayAttack & $73.4 \pm 10.9$ & $51.6 \pm 22.9$ \\
    zugzwang & $77.2 \pm 10.3$ & $70.5 \pm 20.3$ \\
    \midrule
    OVERALL & $70.3 \pm 2.6$ & $53.6 \pm 5.7$ \\
    \bottomrule
  \end{tabular}
\end{table}

KinGPT-Chimera exhibits a statistically significant overall improvement in first-move accuracy (70.3 vs 53.6\%) across all puzzles and is not significantly outperformed by C1-4B on any individual theme. The performance gaps for ``fork'', ``mateIn1'', ``mateIn2'', and ``pin'' themes are especially surprising -- given that these categories of puzzles cover some of the most basic motifs taught to chess beginners. KinGPT models also use much fewer tokens (10 vs. 178 on average) compared to the C1-4B model, showcasing how brevity of generated output does not correlate to more efficient or concise thinking. This suggests that performance on a suite of puzzles is not an accurate standalone metric for measuring language models' chess understanding, as all open-sourced models discussed in this paper are outperformed by a smaller baseline model which is unable to generalize outside the narrow domain of chess puzzles. 

\section{Conclusions}

\subsection{Implications}

The substantial gap in improvement for both move generation accuracy and sanity between pass@10 and modulo-style prompting for general language models, along with the comparable gain in sanity both methods provide for chess-trained language models, suggest several implications.

\begin{enumerate}
    \item{The LLM-Modulo framework is more flexible and less costly for well-defined domains such as chess compared against the methods of pre-training/fine-tuning/post-training on large-scale chess data used by our baseline KinGPT, \cite{feng2023chessgpt}'s ChessGPT, \cite{zhang2025complete}'s ChessLLM, and \cite{tang2026groundedchessreasoninglanguage}'s C1-4B model.}
    \item{Verifier-in-the-loop frameworks recover substantial fractions of performance for general models when comparing against specialized models on domain-specific tasks.}
    \item{Specialized models for well-defined domains can benefit additionally from incorporating a verifier-in-the-loop during inference to provide correctness guarantees for their output.}
\end{enumerate}

\clearpage

\subsection{What about using intermediate thinking traces for interpretability?}
\label{implications-intermediate-tokens}

Another potential use-case of chess-trained language models revolves around interpretability, with \cite{tang2026groundedchessreasoninglanguage} asserting that intermediate thinking traces from models could be used to help explain positions to human users. However, this claim warrants further empirical validation. First, the RLVR loop used by \cite{tang2026groundedchessreasoninglanguage} to verify thinking traces only applies the binary reward by measuring accuracy of the model's final output, without verifying the correctness of the traces themselves. Second, \cite{valmeekam2025beyond} provide an illuminating counter-example, where swapping intermediate tokens for incorrect/incoherent thinking traces yields identical/increased model accuracy despite their semantic meaning being lost, illustrating how increased model performance from intermediate thinking tokens does not guarantee their coherency. Third, \cite{kambhampati2025stop} provide a compelling argument that claiming approximately correct-looking intermediate thinking tokens as valuable interpretability tools for users is both uncalled for and dangerous, with correct-looking yet semantically invalid reasoning traces potentially misleading uninformed users.

\subsection{Future work}

Future work for chess-trained language models should incorporate a verifier-in-the-loop at inference time to guarantee correctness for the model's final output. Evaluating chess-trained language models should be done on a wide variety of chess-related tasks such as in \cite{feng2023chessgpt} and \cite{zhang2025complete}, while ensuring that model training data does not overlap with validation data as done in \cite{tang2026groundedchessreasoninglanguage}. Pass@1 inference should be the norm when evaluating model capabilities across tasks, as pass@10 has proven itself to be an unrealistic and forgiving metric for measuring model performance. In real world settings such as standard chess tournament games, amateur players are not afforded up to 10 tries per move.

Another potential area to explore would be to perform model post-training which also attributes reward to correctness of reasoning traces. For the chess domain, reasoning traces could consist of variation-FEN pairs to track model state coherency across a ``calculated'' variation or utilize the chess position evaluation annotations (+-,+=,=,=-,-+,etc.) used widely in chess literature to measure how models evaluate different positions or variations.

Finally, we outline a hypothetical attempt at creating a chess-trained language model which integrates and improves on the current literature:

\begin{enumerate}
\item{Pre-train a large language model on web-scale corpora, filtering out validation position FENs from its corpus whenever possible. We note that some training/validation overlap of positions will always occur due to opening positions being shared between games; nevertheless, care should be taken to ensure model performance on chess-related benchmarks is not a byproduct of memorization and regurgitation of frequently occurring board states.}
\item{Fine-tune on a wide variety of chess tasks including: move/board state tracking used in \cite{feng2023chessgpt}, playing full games against weak variants of chess engines used in \cite{zhang2025complete}, and solving chess puzzles used in \cite{tang2026groundedchessreasoninglanguage}.}
\item{Utilize RLVR as defined in \cite{tang2026groundedchessreasoninglanguage}, encouraging the use of reasoning traces to increase model performance. As mentioned earlier, validation of intermediate reasoning traces during post-training could also be implemented as an interpretability aid for human users.}
\item{Gate model output for chess-related tasks using the LLM-Modulo framework to provide correctness guarantees for final move legality and accuracy. Besides move validity and move accuracy, additional soft critics could be leveraged to enforce stylistic constraints on a language model's style of moves, such as encouraging aggressive/positional play. Alternatively, implementing hard critics for intermediate reasoning tokens could guarantee correctness at inference time for users utilizing the model as a learning tool.}
\end{enumerate}

\clearpage 

\bibliographystyle{plainnat}  
\bibliography{references}

\newtcolorbox{userbox}{
  colback=blue!5, colframe=blue!50!black,
  title=User, breakable, fonttitle=\bfseries,
}
\newtcolorbox{modelbox}{
  colback=gray!8, colframe=gray!50!black,
  title=Model, breakable, fonttitle=\bfseries,
}
\newtcolorbox{verifierbox}{
  colback=red!4,colframe=red!55!black,  
  title=Verifier, breakable, fonttitle=\bfseries
}

\appendix

\clearpage

\section{LLM-Modulo move validity critic}
\label{modulo_critic1}

The below trace showcases an example of Critic \#1's rejection of an invalid generated move in a given position and its re-prompt with feedback. An image of the current board state is provided for readability; the actual prompts to each model only contained the FEN string of the given position.

\vspace{5pt}

\begin{userbox}
\ttfamily
You are a chess engine. Given the following board position in FEN notation, provide the single best move in UCI format. 

FEN: 2q1nk1r/2r1pp1p/1p1p2p1/1R6/5B2/2Q1P3/5PPP/2R3K1 w - - 0 22

\vspace{5pt}
\begin{center}
    \includegraphics[width=0.6\linewidth]{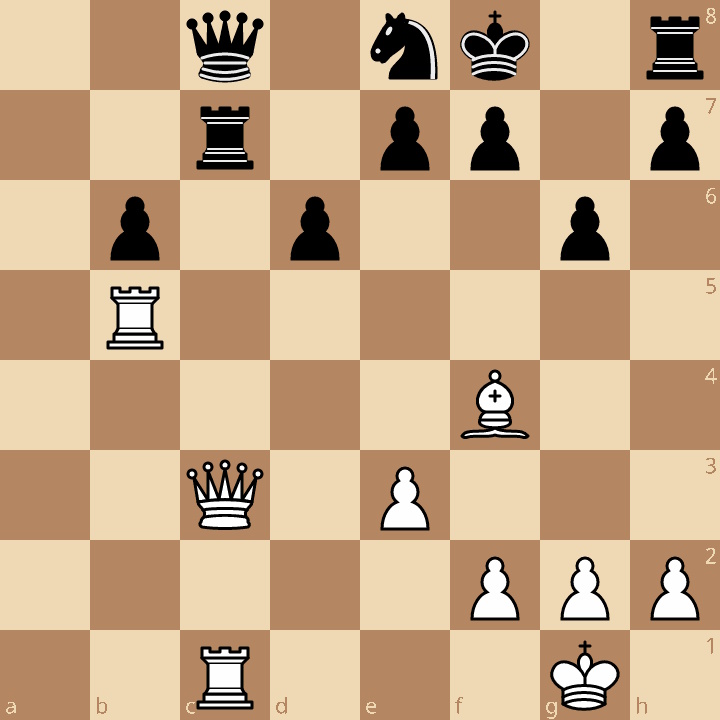}
\end{center}

Best move:
\end{userbox}

\begin{modelbox}
\ttfamily
Rxc7
\end{modelbox}

\begin{verifierbox}
\ttfamily
[INVALID MOVE: CANNOT PARSE VALID UCI/SAN MOVE STRING]
\end{verifierbox}

\begin{userbox}
\ttfamily
The move you provided is invalid. Please provide one of the following legal moves in the position:

b5b6, b5h5, b5g5, b5f5, b5e5, b5d5, b5c5, b5a5, b5b4, b5b3, b5b2, b5b1, f4h6, f4d6, f4g5, f4e5, f4g3, c3h8, c3g7, c3c7, c3f6, c3c6, c3e5, c3c5, c3a5, c3d4, c3c4, c3b4, c3d3, c3b3, c3a3, c3d2, c3c2, c3b2, c3e1, c3a1, g1h1, g1f1, c1c2, c1f1, c1e1, c1d1, c1b1, c1a1, e3e4, h2h3, g2g3, f2f3, h2h4, g2g4
\end{userbox}

\begin{modelbox}
\ttfamily
...
\end{modelbox}

\clearpage

\section{LLM-Modulo move accuracy critic}
\label{modulo_critic2}

The below trace showcases an example of Critic \#2's rejection of a legal but inaccurate move and its re-prompt with feedback. An image of the current board state is provided for readability; the actual prompts to each model only contained the FEN string of the given position.

\vspace{5pt}

\begin{userbox}
\ttfamily
You are a chess engine. Given the following board position in FEN notation, provide the single best move in UCI format.

FEN: 3r4/6Rp/pk6/1p3B2/5N2/P3pbP1/1P5P/4K3 b - - 3 36

\vspace{5pt}
\begin{center}
    \includegraphics[width=0.6\linewidth]{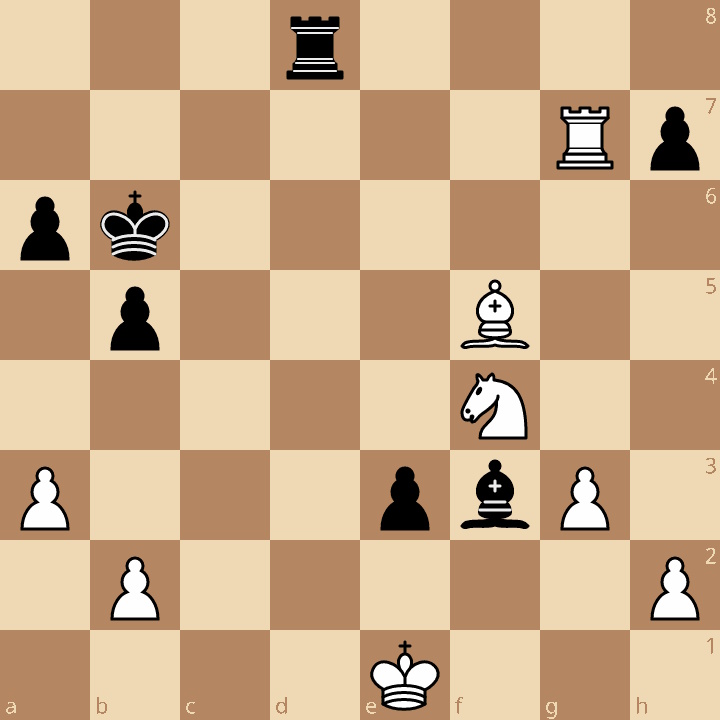}
\end{center}

Best move:
\end{userbox}

\begin{modelbox}
\ttfamily
f3e2
\end{modelbox}

\begin{verifierbox}
\ttfamily
[VALID BUT INACCURATE MOVE: DOES NOT IMPROVE EVALUATION FROM THE ORIGINAL POSITION]
\end{verifierbox}

\begin{userbox}
\ttfamily
The move you provided (f3e2) is valid but does not improve the evaluation of the position. The current position is Mate in 1 for you and the move you provided gives a position with the evaluation of no forced mate. Please try one of the following alternative legal moves instead:

d8h8, d8g8, d8f8, d8e8, d8c8, d8b8, d8a8, d8d7, d8d6, d8d5, d8d4, d8d3, d8d2, d8d1, b6c6, b6c5, b6a5, f3a8, f3b7, f3c6, f3h5, f3d5, f3g4, f3e4, f3g2, f3h1, f3d1, h7h6, a6a5, b5b4, e3e2, h7h5
\end{userbox}

\begin{modelbox}
\ttfamily
d8d1
\end{modelbox}

\begin{verifierbox}
\ttfamily
[MOVE CORRECT]
\end{verifierbox}


\end{document}